\documentclass[10pt,logo]{dobottechreport}

\usepackage[numbers,square,sort&compress]{natbib}

\usepackage{latexsym}

\usepackage{nicefrac}
\usepackage{multirow}
\usepackage{enumitem}
\usepackage{tikz}
\usetikzlibrary{arrows.meta,fit,backgrounds}

\usepackage{placeins}

\newcommand{\method}{MetaNet}
\newcommand{\Method}{MetaNet}
\newcommand{\NA}{\text{--}}


\title{Meta-Learning Where to Allocate Experts:\\
       Task-Conditioned Layer-Wise Compression for MoEs}

\author{%
  \textbf{Wang Rongfeng}$^{1,2,3,*}$ \quad
  \textbf{Shichao Weng}$^{6,*}$ \quad
  \textbf{Zhiqiang Wang}$^{6,*}$ \quad
  \textbf{Xinyu Liu}$^{1}$ \\
  \textbf{Yang Yi}$^{1}$ \quad
  \textbf{Peilong Zhou}$^{1,5}$ \quad
  \textbf{Tang Hongwei}$^{1,2,3,4,\dagger}$ \\
  {\normalfont\Affilfont $^{1}$Institute of Computing Technology, Chinese Academy of Sciences, Beijing, China} \\
  {\normalfont\Affilfont $^{2}$Nanjing College, University of Chinese Academy of Sciences, Nanjing, China} \\
  {\normalfont\Affilfont $^{3}$Nanjing Institute of Information Superbahn, Nanjing, China} \\
  {\normalfont\Affilfont $^{4}$University of Chinese Academy of Sciences, Beijing, China} \\
  {\normalfont\Affilfont $^{5}$School of Advanced Interdisciplinary Sciences, University of Chinese Academy of Sciences, Beijing, China} \\
  {\normalfont\Affilfont $^{6}$Dobot Robotics, Shenzhen, China}
}
\equalcontribution{Equal contribution.}
\projectleader{Corresponding author: tanghongwei@ict.ac.cn}

\begin{document}

\maketitle

\begin{abstract}
Mixture-of-Experts (MoE) models route each token to a subset of expert networks, increasing capacity while keeping per-token computation sparse. In many deployed MoEs, the number of active experts is fixed across layers and tasks, although layer roles and expert redundancy vary with depth~\citep{li2024concept,gao2024higher,qing2024alphalora} and demand varies with difficulty~\citep{huang2024harder,guo2025dynamic,zeng2024adamoe}. Existing approaches address only part of this setting: layer-wise allocations are usually determined offline and reused for all tasks~\citep{gao2024higher,qing2024alphalora}, while token-level methods vary expert activation using local routing signals without task-level context~\citep{huang2024harder,guo2025dynamic,zeng2024adamoe}. We propose MetaNet, a support-set controller that predicts, for each layer, an expert-retention threshold and a bounded routing bias. The backbone, experts, and router remain frozen. On DeepSeek-MoE-16B-Chat, MetaNet provides a tunable accuracy--expert-activation trade-off. Relative to fixed $k{=}6$, a conservative setting activates $3.61$ experts on average ($40\%$ fewer) and achieves comparable MMLU accuracy ($0.489$ vs.\ $0.474$), whereas an aggressive setting activates $2.28$ experts on average ($62\%$ fewer) with accuracy approximately $3.7$ percentage points lower. The MMLU-trained controller also transfers to C-Eval without retraining, activating $2.90$ experts on average ($52\%$ fewer than fixed $k{=}6$) at $0.386$ accuracy.
\end{abstract}
\abscontent

\section{Introduction}

Serving large language models (LLMs) at scale is expensive~\citep{touvron2023llama,openai2023gpt4,ma2023llmpruner}. Mixture-of-Experts (MoE) architectures~\citep{shazeer2017outrageously,fedus2022switch} reduce per-step compute by routing each token to a small subset of expert subnetworks, enabling large total capacity without activating it all at once. This design is used in recent MoE language models such as DeepSeekMoE~\citep{dai2024deepseekmoe} and Mixtral~\citep{jiang2024mixtral}.

Yet standard pretrained MoE inference typically uses the same number of experts at every layer for every token and every task~\citep{shazeer2017outrageously,fedus2022switch,dai2024deepseekmoe}. This contrasts with evidence that layer-wise capacity demands differ. In dense Transformers, layers have distinct functional roles~\citep{li2024concept,tenney2019bert} and contribute differently to task performance~\citep{men2024shortgpt,lad2024robustness}. In MoE models, lower layers exhibit higher expert redundancy than upper layers~\citep{gao2024higher}, and this cross-layer redundancy pattern shifts across tasks~\citep{zhang2026global,guo2025cluster}. Some works exploit depth heterogeneity through layer-wise capacity allocation---OpenELM~\citep{mehta2024openelm} places fewer parameters in shallow layers, while AlphaLoRA~\citep{qing2024alphalora} and LExI~\citep{chittyvenkata2025lexi} assign layer-specific expert budgets---but these allocations are fixed offline and shared across tasks. Token-level dynamic routing~\citep{huang2024harder,guo2025dynamic} lets each token activate a variable number of experts, but decisions are made from per-token routing confidence alone, with no access to task context. Expert pruning~\citep{zhang2026global,guo2025cluster,lu2024notall,yang2024moei2} fixes the retained set at deployment and cannot adapt to novel tasks. Router-modification or construction methods~\citep{hwang2024pregated,do2023hyperrouter} alter the gating mechanism rather than preserving the original router.

This paper studies task-conditioned layer-wise expert allocation for frozen MoE inference. Given a small support set for the current task, \emph{MetaNet} predicts a per-layer expert activation policy: a retention threshold $\rho_{\tau,l}$, which determines how many experts are needed to cover the required routing mass at layer $l$, and a bounded routing bias~$\mathbf{b}_{\tau,l}$, whose magnitude is much smaller than typical gate logits and therefore only weakly perturbs the frozen router. The backbone, original router, and full expert set remain unchanged. On DeepSeek-MoE-16B-Chat, MetaNet reduces the mean number of activated experts by $62\%$ relative to fixed $k{=}6$ (from $6.00$ to $2.28$) with a $3.7$ pp accuracy drop on MMLU, and transfers to C-Eval without retraining with a similar reduction in activated experts. With a weaker budget penalty, it keeps $3.61$ mean experts, achieves comparable accuracy to fixed $k{=}6$, and reduces the mean activated-expert count by $40\%$.

\paragraph{Contributions.}
\begin{itemize}[leftmargin=*, itemsep=1pt]
\item We formulate \emph{task-conditioned per-layer expert allocation} for frozen MoE inference: infer per-layer active expert counts from a small support set, without modifying backbone parameters.
\item We design MetaNet, a small two-branch controller that predicts a retention threshold $\rho_{\tau,l}$ and a bounded routing bias $\mathbf{b}_{\tau,l}$ per layer from support-set routing statistics, keeping the backbone, original router, and full expert pool frozen.
\item We show through component ablations that the budget and prior branches are complementary: the prior branch helps maintain gate-consistent expert selection under tight budgets, and deep layers are more compressible than shallow layers in our setting.
\end{itemize}

\FloatBarrier
\section{Related Work}
\begin{figure}[!htbp]
\centering
\resizebox{\linewidth}{!}{%
\begin{tikzpicture}[
  box/.style={
    draw, rounded corners=2pt,
    minimum width=1.8cm, minimum height=0.65cm,
    font=\small\sffamily, align=center
  },
  token/.style  = {box, fill=gray!12, draw=gray!45},
  router/.style = {box, fill=violet!18, draw=violet!55},
  expact/.style = {box, fill=blue!18, draw=blue!55},
  expoff/.style = {box, fill=gray!8, draw=gray!30, text=gray!40},
  outbox/.style = {box, fill=gray!12, draw=gray!45},
  anarr/.style  = {-{Stealth[length=3.5pt,width=3pt]}, thick},
  actarr/.style = {-{Stealth[length=3.5pt,width=3pt]}, thick, color=blue!65},
  offarr/.style = {dashed, color=gray!35, thin},
]

\node[token]  (tok) at (0,    0)    {Token};
\node[router] (rtr) at (3.2,  0)    {Router};

\node[expoff] (e1)  at (6.4,  2.75) {Expert 1};
\node[expact] (e2)  at (6.4,  1.65) {Expert 2};
\node[expoff] (e3)  at (6.4,  0.55) {Expert 3};
\node[expoff] (e4)  at (6.4, -0.55) {Expert 4};
\node[expact] (e5)  at (6.4, -1.65) {Expert 5};
\node[expoff] (e6)  at (6.4, -2.75) {Expert 6};

\node[outbox] (out) at (9.8,  0)    {Output};

\draw[anarr]  (tok.east) -- (rtr.west);

\draw[offarr] (rtr.east) -- (e1.west);
\draw[offarr] (rtr.east) -- (e3.west);
\draw[offarr] (rtr.east) -- (e4.west);
\draw[offarr] (rtr.east) -- (e6.west);

\draw[actarr] (rtr.east) -- (e2.west);
\draw[actarr] (rtr.east) -- (e5.west);

\draw[actarr] (e2.east) -- (out.west);
\draw[actarr] (e5.east) -- (out.west);

\end{tikzpicture}
}
\caption{An MoE layer: the router activates 2 of 6 experts per token; inactive experts grayed out.}
\label{fig:moe-layer}
\end{figure}
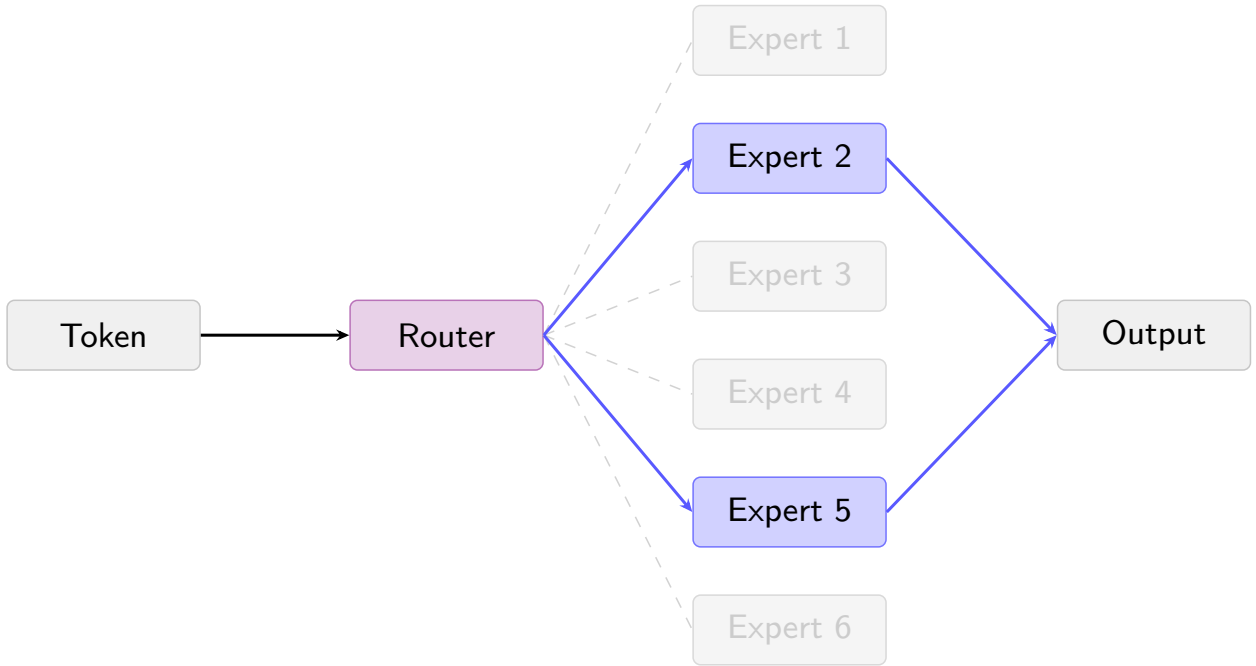

\paragraph{Mixture-of-Experts and inference compression.}
MoE architectures~\citep{shazeer2017outrageously,fedus2022switch} route each input token to $k$ out of $E$ expert FFNs via a gating network (Figure~\ref{fig:moe-layer}), keeping per-step compute proportional to $k$ while allowing large total capacity. DeepSeekMoE~\citep{dai2024deepseekmoe} refines this with fine-grained expert segmentation, and Mixtral~\citep{jiang2024mixtral} shows the effectiveness of sparse expert activation in open LLMs. In these models, however, the active expert count is typically fixed by architecture or inference configuration rather than adapted to the task.

\paragraph{Compressing the inference cost of frozen MoEs.}
Existing work on MoE inference compression can be grouped into four lines. \emph{(i) Expert pruning and merging.} NAEE~\citep{lu2024notall}, MoE-I$^2$~\citep{yang2024moei2}, and STUN~\citep{lee2025stun} prune experts based on calibration-data importance. HC-SMoE~\citep{chen2025hcsmoe} merges experts through hierarchical clustering, whereas ResMoE~\citep{zhang2025resmoe} approximates experts using a Wasserstein barycenter and compressed residuals. These methods determine the compressed expert representation before deployment. \emph{(ii) Input-adaptive computation.} Huang et al.~\citep{huang2024harder}, Dynamic MoE~\citep{guo2025dynamic}, and AdaMoE~\citep{zeng2024adamoe} vary expert activation using token-level signals. Probe Pruning~\citep{le2025probe} is a related batch-wise dynamic structured-pruning method based on model probing rather than MoE router confidence. These approaches make local input-conditioned decisions rather than assigning task-conditioned per-layer capacity from a support set. \emph{(iii) Layer-wise heterogeneous allocation.} MoLA~\citep{gao2024higher}, AlphaLoRA~\citep{qing2024alphalora}, and LExI~\citep{chittyvenkata2025lexi} assign different expert counts to different layers, but the allocation is determined offline and reused across tasks. \emph{(iv) Router modification or construction.} HyperRouter~\citep{do2023hyperrouter} and Pre-gated MoE~\citep{hwang2024pregated} change the routing mechanism, while Read-ME~\citep{cai2024readme} refactorizes a pretrained dense LLM into an MoE with a decoupled pre-gating router. MetaNet instead keeps the original MoE router frozen and adds only a bounded task-conditioned bias.

\paragraph{Systems support for sparse MoE inference.}
Routing sparsity does not by itself guarantee a proportional end-to-end speedup. DeepSpeed-MoE~\citep{rajbhandari2022deepspeedmoe} uses a specialized inference stack, Tutel~\citep{hwang2023tutel} adapts parallelism and pipelining to dynamic expert workloads, and Lina~\citep{li2023lina} addresses all-to-all communication and skewed expert popularity. MetaNet changes the routing policy rather than the runtime and is complementary to these system optimizations.

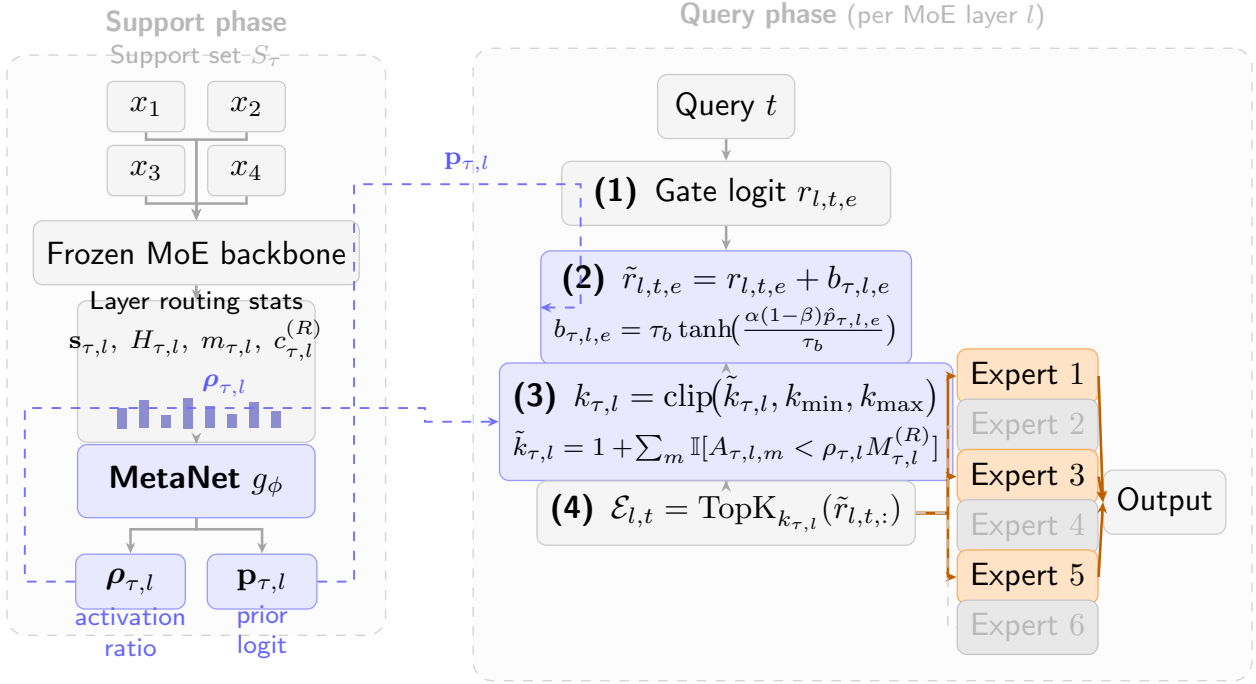
\begin{figure}[!tbp]
\centering
\resizebox{\linewidth}{!}{%
\begin{tikzpicture}[
  font=\normalsize\sffamily,
  block/.style  = {draw, rounded corners=3pt, minimum height=0.70cm,
                   minimum width=3.0cm, align=center, font=\normalsize\sffamily,
                   inner sep=4pt},
  sbox/.style   = {block, fill=gray!8,    draw=gray!40},
  bbox/.style   = {block, fill=blue!9,    draw=blue!40},
  abox/.style   = {block, fill=orange!22, draw=orange!65},
  ibox/.style   = {block, fill=gray!18,   draw=gray!45, text=gray!55},
  snode/.style  = {draw, rounded corners=2pt, fill=gray!8, draw=gray!40,
                   minimum width=0.85cm, minimum height=0.55cm,
                   font=\normalsize\sffamily, align=center},
  obox/.style   = {block, fill=gray!8,    draw=gray!40, minimum width=1.4cm},
  arr/.style    = {-{Stealth[length=3.5pt,width=3pt]}, thick, gray!70},
  aarr/.style   = {-{Stealth[length=3.5pt,width=3pt]}, thick, orange!75!black},
  darr/.style   = {-{Stealth[length=3.5pt,width=3pt]}, dashed,
                   blue!55, semithick},
  sarr/.style   = {dashed, gray!40, thin},
]

\node[snode] (s1) at (-0.55, 3.60) {$x_1$};
\node[snode] (s2) at ( 0.55, 3.60) {$x_2$};
\node[snode] (s3) at (-0.55, 2.90) {$x_3$};
\node[snode] (s4) at ( 0.55, 2.90) {$x_4$};
\node[font=\footnotesize\sffamily, gray!60] at (0, 4.15) {Support set $S_\tau$};

\node[sbox, minimum width=2.6cm] (bb) at (0, 2.00) {Frozen MoE backbone};

\node[sbox, minimum width=2.6cm, minimum height=1.55cm] (stats) at (0, 0.70) {};
\node[font=\footnotesize\sffamily, align=center] at (0, 1.20)
  {Layer routing stats\\[1pt]
   $\mathbf{s}_{\tau,l},\;H_{\tau,l},\;m_{\tau,l},\;c^{(R)}_{\tau,l}$};
\foreach \bx/\bh in {
    -0.82/0.22, -0.58/0.30, -0.34/0.14, -0.10/0.33,
     0.14/0.24,  0.38/0.15,  0.62/0.28,  0.86/0.18}{
  \fill[blue!45!gray, opacity=0.6]
    (\bx-0.04, 0.08) rectangle ++(0.10, \bh);
}

\node[bbox, minimum width=2.6cm, minimum height=0.80cm] (mn) at (0,-0.50)
  {\textbf{MetaNet} $g_\phi$};

\node[bbox, minimum width=1.20cm, minimum height=0.60cm] (rho) at (-0.72,-1.60)
  {$\boldsymbol{\rho}_{\tau,l}$};
\node[font=\footnotesize\sffamily, blue!60, align=center] at (-0.72,-2.15)
  {activation\\ratio};

\node[bbox, minimum width=1.20cm, minimum height=0.60cm] (pp)  at ( 0.72,-1.60)
  {$\mathbf{p}_{\tau,l}$};
\node[font=\footnotesize\sffamily, blue!60, align=center] at ( 0.72,-2.15)
  {prior\\logit};

\foreach \s in {s1,s2,s3,s4}
  \draw[arr] (\s.south) -- ++(0,-0.08) -| (bb.north);
\draw[arr] (bb.south)    -- (stats.north);
\draw[arr] (stats.south) -- (mn.north);
\draw[arr] (mn.south) -- ++(0,-0.12) -| (rho.north);
\draw[arr] (mn.south) -- ++(0,-0.12) -| (pp.north);

\begin{scope}[on background layer]
  \node[draw=gray!35, dashed, rounded corners=6pt, inner sep=8pt,
        fill=gray!2,
        fit=(s1)(s2)(s3)(s4)(bb)(stats)(mn)(rho)(pp),
        label={[font=\small\sffamily, gray!55, yshift=2pt]above:%
               \textbf{Support phase}}] (suppbox) {};
\end{scope}

\node[sbox, minimum width=1.5cm] (qt)   at (5.80, 3.60) {Query $t$};

\node[sbox, minimum width=3.6cm] (gate) at (5.80, 2.65)
  {\textbf{(1)}\enspace Gate logit $r_{l,t,e}$};

\node[bbox, minimum width=3.6cm, minimum height=1.00cm, align=center]
  (padd) at (5.80, 1.40)
  {\textbf{(2)}\enspace$\tilde{r}_{l,t,e}=r_{l,t,e}+b_{\tau,l,e}$\\[2pt]
   \footnotesize$b_{\tau,l,e}=\tau_b\tanh\!\bigl(\tfrac{\alpha(1-\beta)\hat{p}_{\tau,l,e}}{\tau_b}\bigr)$};

\node[bbox, minimum width=3.6cm, minimum height=0.90cm, align=center]
  (bud) at (5.80, 0.15)
  {\textbf{(3)}\enspace$k_{\tau,l}=\mathrm{clip}\!\bigl(\tilde{k}_{\tau,l},k_{\min},k_{\max}\bigr)$\\[2pt]
   \footnotesize$\tilde{k}_{\tau,l}=1+\!\sum_m\mathbb{I}[A_{\tau,l,m}<\rho_{\tau,l}M^{(R)}_{\tau,l}]$};

\node[sbox, minimum width=3.6cm] (topk) at (5.80,-0.85)
  {\textbf{(4)}\enspace$\mathcal{E}_{l,t}=\mathrm{TopK}_{k_{\tau,l}}(\tilde{r}_{l,t,:})$};

\foreach \i/\y/\st in {
    1/ 0.65/abox,  2/ 0.10/ibox,  3/-0.45/abox,
    4/-1.00/ibox,  5/-1.55/abox,  6/-2.10/ibox}{
  \node[\st, minimum width=1.45cm, minimum height=0.55cm]
    (e\i) at (9.10,\y) {Expert $\i$};
}

\node[obox, minimum width=1.3cm] (out) at (10.60,-0.73) {Output};

\draw[arr] (qt.south)   -- (gate.north);
\draw[arr] (gate.south) -- (padd.north);
\draw[arr] (padd.south) -- (bud.north);
\draw[arr] (bud.south)  -- (topk.north);

\foreach \i in {1,3,5}
  \draw[aarr] (topk.east) -- ++(0.35,0) |- (e\i.west);
\foreach \i in {2,4,6}
  \draw[sarr] (topk.east) -- ++(0.35,0) |- (e\i.west);

\foreach \i in {1,3,5}
  \draw[aarr] (e\i.east) -- (out.west);

\begin{scope}[on background layer]
  \node[draw=gray!35, dashed, rounded corners=6pt, inner sep=8pt,
        fill=gray!2,
        fit=(qt)(gate)(padd)(bud)(topk)(e1)(e6)(out),
        label={[font=\small\sffamily, gray!55, yshift=2pt]above:%
               \textbf{Query phase} \footnotesize(per MoE layer $l$)}] {};
\end{scope}


\draw[darr]
  (rho.west) -- ++(-0.55,0)
  -- ++(0, 1.90)
  -- node[above, font=\footnotesize\sffamily, blue!65]{$\boldsymbol{\rho}_{\tau,l}$}
     ++(4.38,0)
  |- (bud.west);

\draw[darr]
  (pp.east) -- ++(0.40,0)
  -- ++(0, 4.35)
  -- node[above, font=\footnotesize\sffamily, blue!65]{$\mathbf{p}_{\tau,l}$}
     ++(2.48,0)
  |- (padd.west);

\end{tikzpicture}
}
\caption{Overview of \method{}. \textbf{Left:} support examples $S_\tau$ pass through the frozen backbone; routing statistics are fed to MetaNet~$g_\phi$, producing a per-layer retention threshold $\rho_{\tau,l}$ and prior score $\mathbf{p}_{\tau,l}$. \textbf{Right:} at query time, the corresponding bounded bias is added to gate logits \textbf{(2)}; cumulative-mass thresholding with $\rho_{\tau,l}$ determines $k_{\tau,l}$ \textbf{(3)}; $\mathrm{TopK}_{k_{\tau,l}}$ selects experts \textbf{(4)}. Orange: activated; gray: skipped. Backbone parameters are unchanged.}
\label{fig:method-overview}
\end{figure}

\paragraph{Meta-learning and task-conditioned controllers.}
MetaNet follows the support-query paradigm~\citep{finn2017maml,vinyals2016matching}: a controller infers a task representation from a small support set and uses it during query-time inference. Unlike MAML~\citep{finn2017maml}, hypernetworks~\citep{ha2017hypernetworks}, or soft-prompt methods~\citep{liu2022fewshot,lester2021prompt}, the controller does not generate or update backbone parameters. It outputs a routing policy---a per-layer retention threshold $\rho_\ell$ and bounded bias $\mathbf{b}_\ell$---that acts on expert selection (Figure~\ref{fig:method-overview}).

\section{Method}

MetaNet derives a per-layer expert-activation policy from a support set while keeping the MoE backbone frozen. Figure~\ref{fig:method-overview} illustrates the full pipeline.

\subsection{Problem setup and support profiling}

Let $f_\theta$ be a pretrained MoE model with $L$ routed layers, each containing $E$ expert FFNs and a router that activates the top $K_{\mathrm{nat}}$ experts per token ($K_{\mathrm{nat}}=6$ for DeepSeek-MoE-16B-Chat, $8$ for OLMoE). Following the support-query paradigm~\citep{finn2017maml,vinyals2016matching}, each task $\tau$ has a support set $S_\tau=\{(x_i,y_i)\}_{i=1}^{N_s}$ (used to infer the task policy) and a query set $Q_\tau$ (used for evaluation).

MetaNet $g_\phi$ maps $S_\tau$ to a per-layer routing policy $\Pi_\tau = g_\phi(S_\tau)$, replacing the fixed top-$K_{\mathrm{nat}}$ with layer-wise activation counts $\mathbf{k}_\tau = (k_{\tau,1},\ldots,k_{\tau,L})$, $k_{\tau,l}\in\{k_{\min},\ldots,k_{\max}\}$. We set $k_{\min}=1$ and $k_{\max}=12$ ($=2K_{\mathrm{nat}}$ for DeepSeek-MoE), so the controller can reduce the budget in redundant layers and retain more experts in layers whose support routing is diffuse.

Let $r_{l,t,e}^{(x)}$ be the gate logit at layer $l$, expert $e$, token $t$ of sample $x$. The per-layer expert score is the token-and-sample average of softmax probabilities:
\begin{equation}
  s_{\tau,l,e}
  = \frac{1}{|S_\tau|}\sum_{x\in S_\tau}
    \frac{1}{T_x}\sum_{t=1}^{T_x}
    \mathrm{softmax}\!\big(r_{l,t}^{(x)}\big)_{\!e},
\end{equation}
giving a distribution $\mathbf{s}_{\tau,l}\in\Delta^{E-1}$ that summarizes which experts the task favors at each layer.

Three scalar statistics capture routing concentration and serve as compressibility signals:
\begin{equation}
\begin{aligned}
  H_{\tau,l}
    &= -\frac{1}{\log E}\sum_{e=1}^{E} s_{\tau,l,e}\log s_{\tau,l,e},\\
  m_{\tau,l}
    &= \max_{e}\; s_{\tau,l,e},
  \qquad
  c^{(R)}_{\tau,l}
    = \sum_{e\,\in\,\mathrm{TopKIdx}_R(\mathbf{s}_{\tau,l})} s_{\tau,l,e}.
\end{aligned}
\end{equation}
Here $H_{\tau,l}\in[0,1]$ is normalized routing entropy ($H=0$: single expert, $H=1$: uniform), $m_{\tau,l}$ is the maximum expert probability, $c^{(R)}_{\tau,l}$ is the fraction of routing mass in the top-$R$ experts, and $\mathrm{TopKIdx}_R$ returns the index set of top-$R$ entries. Layers with low $H_{\tau,l}$, high $m_{\tau,l}$, and high $c^{(R)}_{\tau,l}$ are concentrated and thus amenable to budget reduction.

The support profile uses only quantities produced by the frozen router and does not require gradients through the backbone during policy inference. The same profile is reused for all query examples from the task, so the profiling cost is amortized over the query set. MetaNet therefore acts as a task-level allocator: it chooses how much expert capacity each layer receives, while the original router still ranks experts for individual tokens.

\subsection{MetaNet controller and inference strategy}
\label{subsec:controller}
MetaNet $g_\phi$ processes the per-layer routing statistics through two independent branches and outputs two per-layer signals:
\begin{equation}
  \Pi_\tau
  = g_\phi\!\Big(
      \big\{\mathbf{s}_{\tau,l},H_{\tau,l},m_{\tau,l},c^{(R)}_{\tau,l}\big\}_{l=1}^{L}
    \Big)
  = \big\{\rho_{\tau,l}\!\in\!(0,1],\;
          \mathbf{p}_{\tau,l}\!\in\!\mathbb{R}^{E}\big\}_{l=1}^{L}.
\end{equation}
Both branches share a task encoder: a two-layer MLP that mean-pools $[\mathbf{s}_{\tau,l}; H_{\tau,l}; m_{\tau,l}; c^{(R)}_{\tau,l}]\in\mathbb{R}^{E+3}$ over support examples to produce $\mathbf{z}_\tau\in\mathbb{R}^d$; each layer has a learned embedding $\mathbf{e}_l\in\mathbb{R}^d$ ($d{=}2048$). The \textbf{budget branch} encodes $\mathbf{s}_{\tau,l}$ and the scalar statistics via two-layer MLPs, concatenates with $[\mathbf{z}_\tau;\mathbf{e}_l]$, and produces discrete budget logits whose soft weighted sum gives the \textbf{retention threshold} $\rho_{\tau,l}$. The \textbf{prior branch} uses only the scalar-statistics projection (not $\mathbf{s}_{\tau,l}$, to avoid a trivial copy) concatenated with $[\mathbf{z}_\tau;\mathbf{e}_l]$ to produce a per-expert score $\mathbf{p}_{\tau,l}$, mapped to the \textbf{bounded routing bias} $\mathbf{b}_{\tau,l}$ via scaled $\tanh$.

When multiple experts have similar gate scores, top-$k$ selection is sensitive to small perturbations. The role of $\mathbf{p}_{\tau,l}$ is to provide a weak task-level preference among such experts. We standardize $\mathbf{p}_{\tau,l}$ within each layer ($\hat{p}_{\tau,l,e}=(p_{\tau,l,e}-\mu)/(\sigma+\varepsilon)$) and map it to a bounded routing bias:
\begin{equation}
  b_{\tau,l,e}
    = \tau_b\tanh\!\Big(
        \frac{\alpha(1-\beta)\,\hat{p}_{\tau,l,e}}{\tau_b}
      \Big),
\end{equation}
where $\tau_b>0$ is the maximum bias magnitude, $\alpha\in(0,1)$ is a global scale, and $\beta\in[0,1)$ is a shrinkage coefficient. Because $|b_{\tau,l,e}|\le\tau_b$ and $\tau_b$ is much smaller than typical gate logit magnitudes, the pretrained router remains the main source of expert ordering. In our main configuration $\alpha(1-\beta)=0.005$.

Rather than learning $k_{\tau,l}$ directly (which entangles the budget decision with the shape of the routing distribution), we reformulate budget learning in terms of cumulative probability mass. We combine the support routing distribution with the bounded bias to obtain an adjusted distribution $\bar{\mathbf{q}}_{\tau,l}=\mathrm{softmax}(\log\mathbf{s}_{\tau,l}+\mathbf{b}_{\tau,l})$, used only for budget computation (not for query-phase routing). Here $\log\mathbf{s}_{\tau,l}$ converts the probability vector back to an unnormalized score before adding the bias. Let $M^{(R)}_{\tau,l}=\sum_{e\in\mathrm{TopKIdx}_R(\bar{\mathbf{q}})}\bar{q}_{\tau,l,e}$ be the routing mass covered by the top-$R$ reference experts (here $R=k_{\max}=12$, matching the reference budget). Let $\bar{q}^{\downarrow}_{\tau,l,j}$ denote the $j$-th largest entry of $\bar{\mathbf{q}}_{\tau,l}$ and $A_{\tau,l,m}=\sum_{j=1}^{m}\bar{q}^{\downarrow}_{\tau,l,j}$ its prefix sum. The hard activation count is:
\begin{equation}
  \tilde{k}_{\tau,l}
    = 1 + \sum_{m=1}^{E}
      \mathbb{I}\!\big[A_{\tau,l,m} < \rho_{\tau,l}M^{(R)}_{\tau,l}\big],
  \quad
  k_{\tau,l}
    = \mathrm{clip}\!\big(\tilde{k}_{\tau,l},\,k_{\min},\,k_{\max}\big).
  \label{eq:retention-budget}
\end{equation}
Intuitively, layers with concentrated routing (low $H_{\tau,l}$) reach any given mass threshold with fewer experts and receive a smaller budget. For training, the non-differentiable indicator is replaced by a sigmoid with temperature $T_{\mathrm{bud}}$~\citep{maddison2017concrete,jang2017gumbel,louizos2018l0}:
\begin{equation}
  k_{\tau,l}^{\mathrm{soft}}
    = 1 + \sum_{m=1}^{E}
      \sigma\!\Big(
        \frac{\rho_{\tau,l}M^{(R)}_{\tau,l} - A_{\tau,l,m}}{T_{\mathrm{bud}}}
      \Big).
\end{equation}

At query time, the task-level bias is added to the original gate logit and top-$k_{\tau,l}$ selection is applied:
\begin{equation}
  \tilde{r}_{\tau,l,t,e}
    = r_{l,t,e} + b_{\tau,l,e},
  \qquad
  \mathcal{E}_{\tau,l,t}
    = \mathrm{TopKIdx}_{k_{\tau,l}}\!\big(\tilde{r}_{\tau,l,t,:}\big).
  \label{eq:query-routing}
\end{equation}
The policy $\{\rho_{\tau,l},\mathbf{b}_{\tau,l}\}_{l=1}^{L}$ is computed once from $S_\tau$ and reused across all queries; no backbone parameter is modified.

The formulation separates two decisions. The retention threshold controls \emph{how many} experts are active in a layer, whereas the bounded bias weakly influences \emph{which} experts are preferred when gate scores are close. Fixed top-$k$ baselines keep the first decision constant for all layers and tasks. Token-adaptive methods change $k$ independently for each token, but they do not use a task-level support set. MetaNet keeps the pretrained token router as the fine-grained selector while letting the task profile choose a layer-wise budget before query inference begins.

\subsection{Training objective}
Training follows an episodic support-query protocol~\citep{finn2017maml,vinyals2016matching}: each episode samples a task $\tau$, generates $\Pi_\tau$, and evaluates all loss terms on $Q_\tau$. Only MetaNet parameters $\phi$ are updated. The total loss is:
\begin{equation}
\begin{aligned}
  \mathcal{L}_{\mathrm{total}}
    &= \lambda_{\mathrm{task}}\mathcal{L}_{\mathrm{task}}
    + \lambda_{\mathrm{pp}}\mathcal{L}_{\mathrm{pp}}
    + \lambda_{\mathrm{budget}}\mathcal{L}_{\mathrm{budget}}
    + \lambda_{\mathrm{align}}\mathcal{L}_{\mathrm{align}}\\
    &\quad
    + \lambda_{\mathrm{cons}}\mathcal{L}_{\mathrm{cons}}
    + \lambda_{\mathrm{aux}}\mathcal{L}_{\mathrm{aux}}.
\end{aligned}
\end{equation}

\paragraph{Task loss $\mathcal{L}_{\mathrm{task}}$.} Standard cross-entropy on $Q_\tau$ under the dynamic routing policy of Eq.~\eqref{eq:query-routing}:
\begin{equation}
  \mathcal{L}_{\mathrm{task}}
    = -\frac{1}{|Q_\tau|}\!\sum_{(x,y)\in Q_\tau}\!\log p(y\mid x;\,\Pi_\tau).
\end{equation}

\paragraph{Performance-preserving penalty $\mathcal{L}_{\mathrm{pp}}$.} We compute a detached reference loss $\mathcal{L}_{\mathrm{ref}}$ by evaluating $Q_\tau$ under the original router at native top-$K_{\mathrm{nat}}$. Let $\Delta = \mathcal{L}_{\mathrm{task}} - \mathcal{L}_{\mathrm{ref}} - m_{\mathrm{pp}}$, where $m_{\mathrm{pp}}\ge0$ is a tolerance margin. The penalty is a one-sided Huber loss: $\mathcal{L}_{\mathrm{pp}}=0$ when $\Delta\le0$, and
\begin{equation}
  \mathcal{L}_{\mathrm{pp}} =
  \begin{cases}
    \Delta^2/(2\delta_H) & \text{if } 0 < \Delta \le \delta_H, \\
    \Delta - \delta_H/2  & \text{if } \Delta > \delta_H,
  \end{cases}
\end{equation}
where $\delta_H>0$ is the Huber transition point (quadratic below $\delta_H$, linear above).

\paragraph{Budget loss $\mathcal{L}_{\mathrm{budget}}$.} Let $\bar{k}_{\tau,l}=(k_{\tau,l}^{\mathrm{soft}}-k_{\min})/(k_{\max}-k_{\min})$ be the normalized soft budget, $u_{\tau,l}=u_0+(1-u_0)H_{\tau,l}$ be the per-layer compression target (high-entropy layers get a relaxed target; $u_0=0.2$), and $s_{\mathrm{safe}}=\sigma(-\Delta/T_q)$ be a soft safety gate (with temperature $T_q>0$) that suppresses compression pressure when quality degrades. The budget loss combines compression and rescue objectives:
\begin{equation}
\begin{aligned}
  \mathcal{L}_{\mathrm{budget}}
    &= w_{\mathrm{cmp}}\,s_{\mathrm{safe}}
       \!\Big[
         \tfrac{1}{L}\sum_l\mathrm{ReLU}(\bar{k}_{\tau,l}-u_{\tau,l})
         + w_{\mathrm{compute}}\tfrac{1}{L}\sum_l\bar{k}_{\tau,l}
       \Big] \\
    &\quad
     + w_{\mathrm{rescue}}\,(1-s_{\mathrm{safe}})
       \cdot \tfrac{1}{L}\sum_l\mathrm{ReLU}(u_{\tau,l}-\bar{k}_{\tau,l}).
\end{aligned}
\end{equation}
When quality is safe ($s_{\mathrm{safe}}\approx1$) the compression term dominates; when quality degrades ($s_{\mathrm{safe}}\approx0$) the rescue term takes over and increases the budget.

\paragraph{Alignment and consistency losses.} $\mathcal{L}_{\mathrm{align}}$ constrains the routing bias to remain anchored to the support routing distribution, preventing the prior branch from inverting the original gate ordering: $\mathcal{L}_{\mathrm{align}}=\frac{1}{L}\sum_l\mathrm{KL}(\mathbf{s}_{\tau,l}\|\mathrm{softmax}(\mathbf{b}_{\tau,l}))$, where the KL is computed with the support distribution as the target so that the learned bias is penalized for deviating from it. $\mathcal{L}_{\mathrm{cons}}$ encourages policy stability: a second policy $(\rho_{\tau,l}',\mathbf{p}_{\tau,l}')$ is computed from a random half-split of $S_\tau$, and the loss penalizes disagreement with the full-set policy: $\mathcal{L}_{\mathrm{cons}}=\frac{1}{L}\sum_l[(\rho_{\tau,l}-\rho_{\tau,l}')^2+\|\mathbf{p}_{\tau,l}-\mathbf{p}_{\tau,l}'\|^2]$.

\paragraph{Auxiliary regularizers $\mathcal{L}_{\mathrm{aux}}$.} Three small-weight terms stabilize training: a budget-diversity term $\mathcal{L}_{\mathrm{res}}=\frac{1}{L}\sum_l(\bar{k}_{\tau,l}-\bar{u})^2$ (suppresses constant-budget solutions); a negative-entropy reward $\mathcal{L}_{\mathrm{ent}}$ on $\mathrm{softmax}(\mathbf{p}_{\tau,l})$ (prevents prior collapse); and a rank loss $\mathcal{L}_{\mathrm{rank}}$ that enforces soft monotonicity between routing entropy and budget:
\begin{equation}
  \mathcal{L}_{\mathrm{rank}}
    = \frac{1}{|\mathcal{P}|}\sum_{(l,l')\in\mathcal{P}}
      \mathrm{ReLU}\!\big(\bar{k}_{\tau,l} - \bar{k}_{\tau,l'} + m_{\mathrm{r}}\big),
\end{equation}
where $\mathcal{P}=\{(l,l')\,:\,H_{\tau,l}<H_{\tau,l'}\}$ and $m_{\mathrm{r}}=0.05$. Full hyperparameter values are in Table~\ref{tab:hyperparams}.

\FloatBarrier
\section{Experimental setup}
\label{sec:experimental-setup}

\subsection{Models and evaluation}

The main backbone is DeepSeek-MoE-16B-Chat~\citep{dai2024deepseekmoe} ($27$ routed MoE layers, $64$ sparse experts per layer, native top-$k$ $K_{\mathrm{nat}}{=}6$); all backbone parameters are frozen and only MetaNet $g_\phi$ is updated. Experiments run on three NVIDIA RTX~4090 24\,GB GPUs (bfloat16, \texttt{device\_map=auto}). MMLU episodes use $N_s=8$ support and $N_q=16$ disjoint query examples. For C-Eval, all five labeled development examples per subject form the support set and up to $16$ labeled validation examples form the query set. MetaNet is trained for $450$ episodic steps on MMLU~\citep{hendrycks2021mmlu} meta-train subjects; the checkpoint with highest meta-validation accuracy is selected and evaluated on held-out meta-test subjects.

We run four experiments: \textbf{(1) Main}---MMLU~\citep{hendrycks2021mmlu} meta-test ($12$ subtasks) vs.\ fixed top-$k$ and HyperRouter-style; C-Eval~\citep{huang2023ceval} as zero-shot transfer. \textbf{(2) Cross-backbone}---the MMLU-trained checkpoint applied zero-shot to OLMoE-1B-7B~\citep{muennighoff2024olmoe}. \textbf{(3) Cross-architecture}---GoogLeNet~\citep{szegedy2015going} (frozen) as backbone on two synthetic episodic vision tasks (\texttt{syn-pattern}/\texttt{syn-pattern-shift}), selecting a $3$-of-$4$ branch policy (active ratio $0.75$). \textbf{(4) Ablations}---component (budget-only, prior-only, full) and layer-position (shallow/middle/deep/all).

\textbf{Synthetic vision episodes.} \texttt{syn-pattern} and \texttt{syn-pattern-shift} are procedural episodic tasks used only as a cross-architecture check. Each class is generated from a simple visual prototype, including color blocks, vertical or horizontal stripes, diagonal patterns, and ring-like structures, with random noise added to each image. The shifted split, \texttt{syn-pattern-shift}, additionally applies horizontal flipping and mild color-statistic perturbation. Within each episode, support and query images are sampled from the same synthetic class distribution. The support set is used to infer the branch policy, and the query set evaluates whether this policy transfers to unseen samples from the same episode. This experiment tests whether support-conditioned structural control can be applied outside MoE routing.

For MMLU~\citep{hendrycks2021mmlu} and C-Eval~\citep{huang2023ceval}, each subject is treated as a task. The fixed MMLU split contains $34$ meta-train, $11$ meta-validation, and $12$ meta-test subjects. Subject identities are disjoint across the three partitions, so the main MMLU result contains $12\times16=192$ held-out query predictions rather than additional samples from training subjects. Within each episode, support examples are used only to construct the routing profile and query examples are used for the reported metric.

\paragraph{Evaluation scopes.}
We distinguish the main transfer evaluation from the additional analyses. \textsc{C-Eval-Full} applies the MMLU-trained controller to all $52$ C-Eval subjects without parameter updates and contains $828$ labeled query examples. The expanded baselines and operating-point sweep use a fixed $10$-subject subset with $16$ query examples per subject ($160$ in total), denoted \textsc{C-Eval-10}. Results from \textsc{C-Eval-Full} and \textsc{C-Eval-10} are reported in separate panels and are not compared directly. The ten subjects are listed in Appendix~\ref{sec:additional-details}.

\paragraph{Sampling seeds.}
Unless stated otherwise, results use one episode per subject. In the repeated-sampling check, the subject split remains fixed and the seed changes only support/query example sampling; support and query examples remain disjoint. Two seeds and two episodes per held-out MMLU subject yield $384$ query predictions per seed. Fixed $k{=}6$, fixed $k{=}12$, and both MetaNet profiles are evaluated under both seeds. Appendix~\ref{subsec:sampling-sensitivity} reports each run and the mean with sample standard deviation.

\subsection{Baselines and metrics}

\textbf{Baselines.} Fixed top-$k$ uses $k\in\{1,\ldots,6,12\}$; $k{=}6$ is the native router budget and $k{=}12$ is an upper-budget reference rather than a compute-matched baseline. HyperRouter-style~\citep{do2023hyperrouter} uses a linear head over the same support statistics while keeping top-$k$ fixed at $6$. The expanded comparison also includes inference-compatible adaptations inspired by AdaMoE~\citep{zeng2024adamoe}, Dynamic MoE~\citep{guo2025dynamic}, and Probe Pruning~\citep{le2025probe}. The rows are labeled ``adapted'' because they place the corresponding decision rule in the same frozen DeepSeek harness rather than reproduce each method's original architecture and training procedure.

\textbf{Metrics.} For each task $\tau$, accuracy is the exact-match rate on the query set $Q_\tau$; reported accuracies average this value over evaluation tasks. The mean dynamic top-$k$ is $\bar{k}=\frac{1}{|\mathcal{T}|L}\sum_{\tau\in\mathcal{T}}\sum_{l=1}^{L}k_{\tau,l}$, where $k_{\tau,l}$ is the layer budget selected from the support set. The active expert ratio is $\bar{k}/E$ and serves as a routed-expert workload proxy rather than a measurement of total model FLOPs; for a fixed top-$k$ baseline it reduces to $k/E$ (e.g., $6/64=0.094$). Latency is the measured wall-clock time per query in the common three-GPU harness used for the additional experiments. It characterizes the current implementation rather than an optimized MoE serving stack and is reported separately from active ratio because dispatch, dense layers, device communication, and runtime overhead are not represented by $\bar{k}/E$. Ablations additionally report \emph{gate agreement}, the Jaccard similarity between the expert set selected with MetaNet's biased gate and the set selected by the original unbiased gate, averaged over layers and query examples, and \emph{low-use selection}, the fraction of activated experts whose support-set routing mass lies in the bottom quartile of all experts at the same layer.

\FloatBarrier
\section{Results}

\subsection{Main results}

\begin{table*}[!t]
\centering
\footnotesize
\setlength{\tabcolsep}{4pt}
\begin{tabular*}{0.78\textwidth}{@{\extracolsep{\fill}}lrrr@{}}
\toprule
Method & Accuracy & Act.\ ratio & Mean $k$ \\
\midrule
\multicolumn{4}{l}{\textbf{MMLU (meta-test, 12 tasks $\times$ 16 queries)}} \\
Fixed $k{=}6$           & 0.474 & 0.094 & 6.00 \\
Fixed $k{=}12$          & 0.495 & 0.188 & 12.00 \\
Fixed $k{=}4$           & 0.432 & 0.063 & 4.00 \\
Fixed $k{=}3$           & 0.417 & 0.047 & 3.00 \\
HyperRouter-style       & 0.406 & 0.094 & 6.00 \\
\method{} (aggressive)  & 0.438 & \textbf{0.036} & \textbf{2.28} \\
\method{} (conservative)& \textbf{0.489} & 0.056 & 3.61 \\
\midrule
\multicolumn{4}{l}{\textbf{C-Eval-Full (52 subjects, 828 queries; zero-shot transfer)}} \\
Fixed $k{=}6$           & 0.444 & 0.094 & 6.00 \\
Fixed $k{=}2$           & 0.377 & \textbf{0.031} & \textbf{2.00} \\
Fixed $k{=}4$           & 0.443 & 0.063 & 4.00 \\
\method{} (aggressive)  & 0.386 & 0.045 & 2.90 \\
\method{} (conservative)& \textbf{0.452} & 0.056 & 3.61 \\
\bottomrule
\end{tabular*}
\caption{Main results on MMLU and C-Eval-Full. Bold accuracy marks the stronger MetaNet setting at or below the native $k{=}6$ budget; bold active-ratio and mean-$k$ values mark the minimum routed-expert workload in each panel. In the MMLU panel, fixed $k{=}12$ is an upper-budget reference and is excluded from the bold accuracy comparison. The C-Eval fixed-$k{=}12$ result was measured only on C-Eval-10 and is reported separately in Table~\ref{tab:expanded-baselines}. The conservative setting uses a lower budget penalty weight $\lambda_{\mathrm{budget}}$.}
\label{tab:main-results}
\end{table*}

Table~\ref{tab:main-results} summarizes MMLU and C-Eval-Full results. \textbf{Compression vs.\ accuracy.} \Method{} reaches accuracy $0.438$ at mean top-$k$ $2.28$ (ratio $0.036$, $62\%$ fewer activated experts than fixed $k{=}6$); fixed $k{=}3$ has lower accuracy ($0.417$) and a larger active ratio ($0.047$), indicating that a uniform budget does not match the layer-wise allocation learned by MetaNet. \textbf{Effect of support-conditioned routing.} The HyperRouter-style baseline, matched to ratio $0.094$, loses $6.8$ points relative to fixed $k{=}6$ ($0.406$ vs.\ $0.474$), showing that support conditioning alone is insufficient here without budget control and a bounded interaction with the frozen gate. \textbf{Transfer to C-Eval.} The MMLU-trained controller transfers to C-Eval-Full without retraining ($2.90$ mean experts, ratio $0.045$, $52\%$ fewer activated experts than fixed $k{=}6$), outperforming the closest fixed baseline ($k{=}2$, ratio $0.031$, acc.\ $0.377$) at a nearby expert budget.

Table~\ref{tab:external} reports cross-backbone transfer. Applied to OLMoE-1B-7B~\citep{muennighoff2024olmoe} without updating weights, \method{} exceeds fixed $k{=}8$ on both MMLU ($0.573$ vs.\ $0.521$) and C-Eval ($0.375$ vs.\ $0.363$) with mean top-$k{=}7.52$. Since this budget is close to OLMoE's native $k{=}8$, the improvement may reflect the bounded routing bias rather than expert-count reduction alone (Section~\ref{subsec:mechanistic}). Table~\ref{tab:exp3-transfer} gives the GoogLeNet cross-architecture check: \method{} reaches $0.66$ on \texttt{syn-pattern-shift} vs.\ $0.58$ for static $3/4$-branch selection at the same active ratio.

The conservative and aggressive settings show how the objective controls the operating point. The aggressive objective favors smaller budgets and indicates that many layers can operate below the native $k{=}6$ budget, at the cost of some accuracy. The conservative objective lowers the compression pressure and keeps more experts active when the quality guard predicts risk; this reduces the MMLU gap while still lowering the average active-expert count from $6.0$ to $3.61$. The two settings use the same controller family under different deployment budgets.

\subsection{Expanded baselines and operating points}

Table~\ref{tab:expanded-baselines} reports the additional comparisons under a common frozen-inference harness and restores the fixed $k{=}12$ reference in the C-Eval-10 panel. The MMLU panel retains the $12$-subject, $192$-query scope, whereas the C-Eval panel uses C-Eval-10 ($10$ subjects, $160$ queries). The AdaMoE and Dynamic MoE adaptations remain close to six experts on average. MetaNet provides operating points with lower routed-expert workload; the Probe Pruning adaptation is faster but loses substantial accuracy. Section~\ref{sec:experimental-setup} states the different deployment assumptions.

Table~\ref{tab:operating-points} gives the denser operating-point sweep. The tags \texttt{bt0p3}, \texttt{bt0p4}, \texttt{bt0p5}, and \texttt{bt0p8} encode nominal base budget targets $u_0\in\{0.3,0.4,0.5,0.8\}$ for independently trained controllers in the same family; they are not achieved active ratios. The resulting points are not strictly monotonic, so we report the achieved mean top-$k$ and active ratio for every run rather than describe every candidate as Pareto-optimal. Appendix~\ref{subsec:sampling-sensitivity} reports matched seed-$0$/seed-$1$ runs. Conservative MetaNet obtains $0.506\pm0.022$ accuracy, and aggressive MetaNet obtains $0.477\pm0.022$ (mean $\pm$ sample standard deviation).

\begin{table}[!tbp]
\centering
\footnotesize
\setlength{\tabcolsep}{1pt}
\begin{tabular*}{\linewidth}{@{\extracolsep{\fill}}llrrr@{}}
\toprule
Backbone & Method & MMLU & C-Eval & Mean $k$ \\
\midrule
\multirow{4}{*}{\shortstack[l]{DeepSeek-MoE\\16B-Chat\\($K_\mathrm{nat}{=}6$, 64 exp.)}}
 & Fixed $k{=}6$     & 0.474 & 0.444 & 6.00 \\
 & Fixed $k{=}4$     & 0.432 & 0.443 & 4.00 \\
 & Fixed $k{=}2$     & 0.365 & 0.377 & 2.00 \\
 & \method{}         & 0.438 & 0.386 & 2.28 \\
\midrule
\multirow{4}{*}{\shortstack[l]{OLMoE-1B-7B\\($K_\mathrm{nat}{=}8$, 64 exp.)}}
 & Fixed $k{=}8$     & 0.521 & 0.363 & 8.00 \\
 & Fixed $k{=}4$     & 0.479 & 0.350 & 4.00 \\
 & Fixed $k{=}2$     & 0.417 & 0.350 & 2.00 \\
 & \method{}         & \textbf{0.573} & \textbf{0.375} & 7.52 \\
\bottomrule
\end{tabular*}
\caption{Cross-backbone generalization (Exp.~II). MetaNet trained on DeepSeek and evaluated zero-shot on OLMoE-1B-7B~\citep{muennighoff2024olmoe}. Fixed-$k$ rows report each backbone's native and lower budgets; the final column reports mean $k$ on MMLU.}
\label{tab:external}
\end{table}

\begin{table}[!tbp]
\centering
\footnotesize
\setlength{\tabcolsep}{2.5pt}
\begin{tabular*}{\linewidth}{@{\extracolsep{\fill}}llrr@{}}
\toprule
Dataset & Method & Accuracy & Act.\ ratio \\
\midrule
\multirow{3}{*}{\texttt{syn-pattern}} & Original & \textbf{1.00} & 1.00 \\
 & Static   & \textbf{1.00} & 0.75 \\
 & \method{}& 0.98          & 0.75 \\
\midrule
\multirow{3}{*}{\texttt{syn-pattern-shift}} & Original & \textbf{0.99} & 1.00 \\
 & Static   & 0.58          & 0.75 \\
 & \method{}& \textbf{0.66} & 0.75 \\
\bottomrule
\end{tabular*}
\caption{Cross-architecture proof-of-concept results (Experiment~III, GoogLeNet backbone).}
\label{tab:exp3-transfer}
\end{table}

\subsection{Ablation studies}

\paragraph{Method components.}
Table~\ref{tab:method-ablation} ablates the two branches. Removing the prior branch (Budget-only) drops accuracy to $0.375$ and gate agreement to $0.038$: the budget branch alone selects expert sets with little overlap with the original gate. Removing the budget branch (Prior-only) uses $k{=}12$ throughout and recovers the highest accuracy ($0.484$), but at the full $k{=}12$ routing budget. With both branches, gate agreement rises to $0.943$ and low-use selection drops to $0.004$, indicating that the prior branch keeps expert selection close to the original gate while the budget branch reduces active experts.

\begin{table}[!tbp]
\centering
\footnotesize
\setlength{\tabcolsep}{1.5pt}
\begin{tabular*}{\linewidth}{@{\extracolsep{\fill}}lrrrrr@{}}
\toprule
Variant & Acc. & Act.\ ratio & Mean $k$ & Gate agr. & Low-use \\
\midrule
Fixed $k{=}6$   & 0.474 & 0.094 & 6.00 & \NA   & \NA   \\
Budget-only     & 0.375 & 0.045 & 2.89 & 0.038 & 0.734 \\
Prior-only      & \textbf{0.484} & 0.188 & 12.00 & 0.622 & 0.318 \\
Full method     & 0.438 & \textbf{0.036} & \textbf{2.28} & \textbf{0.943} & \textbf{0.004} \\
\bottomrule
\end{tabular*}
\caption{Method-component ablation on MMLU. Dashes denote controller diagnostics omitted for fixed routing.}
\label{tab:method-ablation}
\end{table}

\paragraph{Layer position.}
Table~\ref{tab:layer-mask} restricts compression to different layer ranges, holding all other layers at fixed $k{=}6$. Compressing only the deep layers (layers $18$--$26$) costs $0.005$ accuracy points relative to the uncompressed baseline while reaching ratio $0.069$, the best efficiency--accuracy trade-off in this sweep. Compressing only the shallow layers (layers $0$--$8$) loses $0.078$ accuracy points at ratio $0.087$, indicating that early-layer expert selection is less redundant. Middle, FrontBack-10, and Random-9 configurations all reach $0.427$. Full-layer compression pushes ratio down to $0.036$ and achieves $0.438$ accuracy, showing that joint optimization across all layers can partially compensate for the harder shallow-layer budget.

\begin{table}[!tbp]
\centering
\footnotesize
\setlength{\tabcolsep}{4pt}
\begin{tabular*}{\linewidth}{@{\extracolsep{\fill}}lrrr@{}}
\toprule
Variant & Acc. & Act.\ ratio & Mean $k$ \\
\midrule
Fixed $k{=}6$ (all layers) & 0.474 & 0.094 & 6.00 \\
All layers             & 0.438 & 0.036 & 2.28 \\
Shallow-9 (0--8)       & 0.396 & 0.087 & 5.59 \\
Middle-9               & 0.427 & 0.074 & 4.75 \\
Deep-9 (18--26)        & \textbf{0.469} & \textbf{0.069} & \textbf{4.40} \\
FrontBack-10           & 0.427 & 0.081 & 5.16 \\
Random-9               & 0.427 & 0.077 & 4.94 \\
\bottomrule
\end{tabular*}
\caption{Layer-mask ablation on MMLU.}
\label{tab:layer-mask}
\end{table}

\begin{figure*}[!t]
\centering
\includegraphics[width=0.70\textwidth,trim=0 0 0 58,clip]{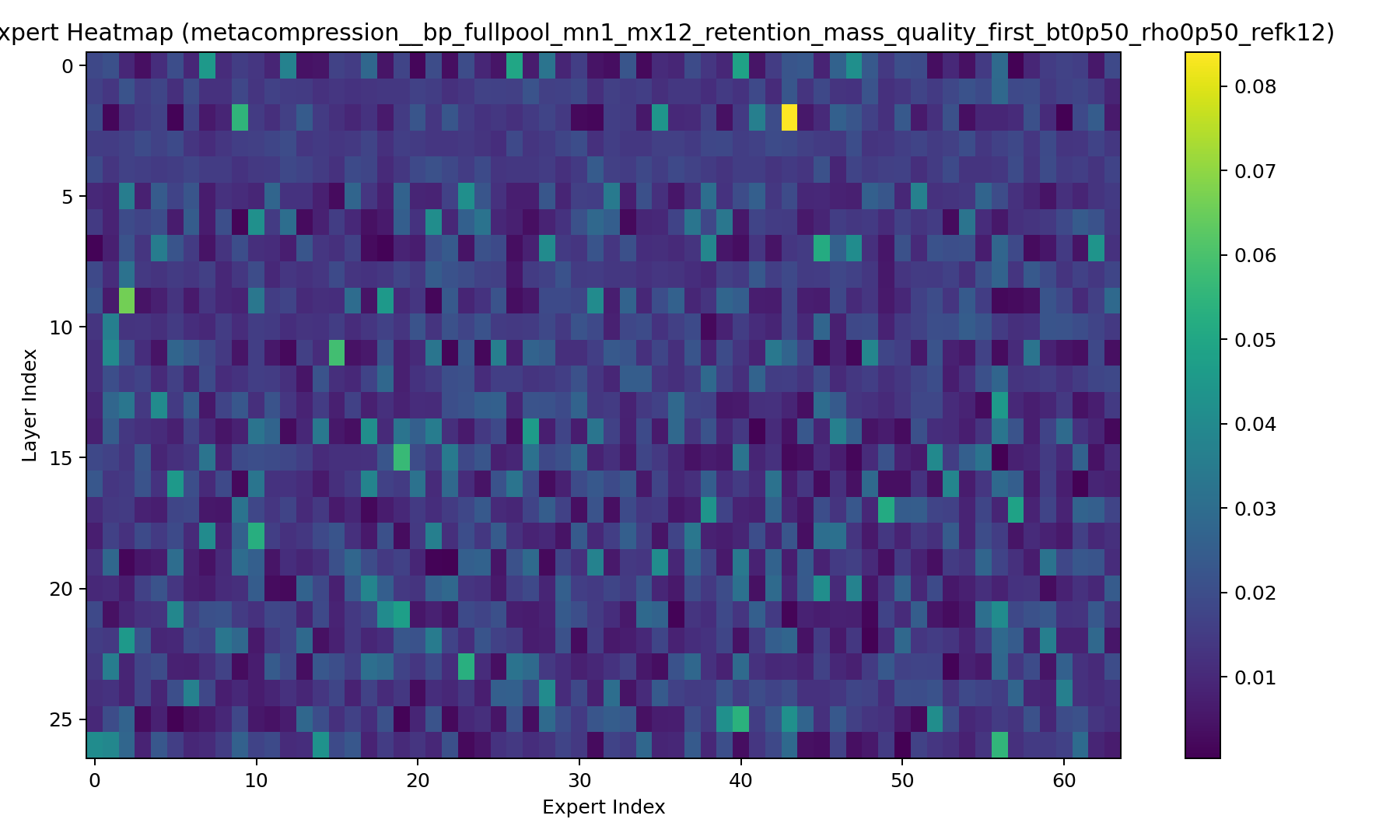}
\caption{Support-derived layer--expert routing profile for the selected MMLU checkpoint ($27$ routed layers $\times$ $64$ experts; color denotes mean normalized routing mass).}
\label{fig:layer-heatmap}
\end{figure*}

\subsection{Mechanistic analysis}
\label{subsec:mechanistic}

Figure~\ref{fig:layer-heatmap} visualizes the support-derived routing mass over $27$ layers and $64$ experts. Expert use changes substantially with depth rather than following one layer-invariant profile. The mean cross-layer Jaccard overlap of activated expert sets is $0.10$ (max $0.26$), computed by averaging pairwise Jaccard similarity between layer-level activated expert sets within each task; low overlap indicates that different layers often select different expert subsets. Both latency harnesses show modest reductions. In the original harness, MetaNet runs in $491.1$\,ms per query vs.\ $508.5$\,ms for fixed $k{=}6$ ($3.4\%$ lower). In the separate common three-GPU harness, aggressive MetaNet reduces latency from $542.94$ to $516.52$\,ms ($4.9\%$) while reducing mean top-$k$ from $6.00$ to $2.28$ ($62.0\%$ fewer activated experts). Measurements from the two harnesses are not pooled. The gap between the reductions in mean top-$k$ and latency indicates that the current inference stack does not convert routing sparsity proportionally into end-to-end speedup: MetaNet changes routed-expert execution, whereas dense attention, token dispatch and permutation, synchronization, and inter-device communication remain. This distinction is consistent with system-level MoE work that introduces specialized kernels, adaptive execution, or communication scheduling~\citep{rajbhandari2022deepspeedmoe,hwang2023tutel,li2023lina}. The support-profiling pass is computed once per episode and reused across all queries. On OLMoE, \method{} exceeds fixed $k{=}8$ by $5.2$ points ($0.573$ vs.\ $0.521$); because its mean top-$k$ is close to the native $k{=}8$, this experiment supports cross-backbone transfer but provides only weak evidence for compression. The bounded bias is a plausible contributor, but the experiment does not isolate that mechanism.

The nonuniform support profile helps explain why one fixed top-$k$ can be inefficient, while the layer-position ablation supplies the direct budget evidence: deep-layer compression is relatively safe, but optimizing all layers jointly gives a more compressed operating point. MetaNet uses the task support set to construct this layer-wise policy rather than relying on one handcrafted layer mask.

\FloatBarrier
\section{Conclusion}

MetaNet is a support-set controller for frozen MoE inference. It predicts per-layer expert budgets and a bounded routing bias without updating backbone parameters. Across MMLU, C-Eval, a second MoE backbone, and a vision backbone, the results show that expert demand varies by task and depth. Modeling this variation gives better accuracy--expert-activation trade-offs than uniform fixed-$k$ routing.

\section*{Limitations}

MetaNet reduces activated-expert workload rather than model storage or total end-to-end computation. The active ratio is a routing-level proxy, not a measurement of total model FLOPs, and the full expert pool remains resident in memory. The method therefore does not provide proportional parameter-storage or peak-memory savings in the current implementation. Wall-clock speedup depends on whether the runtime can exploit the smaller routed-expert set; without specialized dispatch and communication support, latency need not decrease in proportion to mean top-$k$. Integrating MetaNet with optimized MoE runtimes is left for future work.

The main MMLU result uses $192$ held-out queries, and the repeated-sampling check contains only two seeds. We report both runs and their sample standard deviation for the fixed-budget and MetaNet profiles, but two seeds do not support a reliable confidence interval or significance test. C-Eval-Full and C-Eval-10 have different scopes and are reported separately. The adapted baselines share our frozen inference harness but are not exact reproductions of methods that require different training or architectures. In addition, the controller is evaluated mainly on academic multiple-choice benchmarks; open-ended generation, long-context tasks, and safety-sensitive downstream applications may exhibit different routing patterns. Finally, the best layer-wise budget location is treated as a learned outcome rather than a fully characterized principle.

\section*{Ethical Considerations}

This work studies inference-time efficiency for released MoE models and does not introduce a new pretrained language model, dataset, or user-facing deployment system. Lower activated-expert workload can reduce hardware cost and energy use when paired with suitable sparse dispatch kernels. Potential risks are indirect: cheaper inference may make existing LLM misuse scenarios easier to scale. The method does not add new generative capabilities, collect personal data, or rely on human-subject experiments. We use publicly available research benchmarks and model checkpoints according to their release terms, report aggregate evaluation metrics only, and will release code and controller artifacts for research reproducibility.

\newpage
\section*{Acknowledgments}

This work was supported by the Major Science and Technology Project of Henan Province, China, entitled \emph{Research and Industrialization of Key Technologies for Intelligent Computing in Large-Scale Video Scenarios under Digital Social Governance} (Grant No.~241100210100).

\FloatBarrier
\small
\bibliographystyle{unsrtnat}
\bibliography{references}

\normalsize
\clearpage
\appendix

\section{Hyperparameters}
\begin{table}[!htbp]
\centering
\footnotesize
\setlength{\tabcolsep}{4pt}
\begin{tabular*}{\linewidth}{@{\extracolsep{\fill}}ll@{}}
\toprule
Parameter & Value \\
\midrule
\multicolumn{2}{l}{\textit{Backbone \& modules}} \\
Model & \texttt{DeepSeek-MoE-16B-Chat} \\
Frozen & backbone, experts, original gate \\
Trainable & MetaNet only \\
Routed experts / layer; native top-$k$ ($K_{\mathrm{nat}}$) & 64;\enspace 6 \\
\midrule
\multicolumn{2}{l}{\textit{Retention profile}} \\
Reference top-$k$ ($R$); routing strategy & 12;\enspace \texttt{retention\_mass} \\
$k_{\min}$;\enspace $k_{\max}$ & 1;\enspace 12 \\
\midrule
\multicolumn{2}{l}{\textit{MetaNet training}} \\
Steps; learning rate & 450;\enspace $5\times10^{-4}$ \\
\midrule
\multicolumn{2}{l}{\textit{Weak prior}} \\
$\alpha$;\enspace $\beta$;\enspace max bias $\tau_b$ & 0.05;\enspace 0.90;\enspace 0.05 \\
\midrule
\multicolumn{2}{l}{\textit{Objective}} \\
Budget loss mode & \texttt{quality\_first} \\
Rescue $w_{\mathrm{rescue}}$; compute $w_{\mathrm{compute}}$ & 6.0;\enspace 0.12 \\
Task scale; layer scale & 0.12;\enspace 0.18 \\
Overshoot weight; upper tolerance & 2.5;\enspace 0.05 \\
Base compression target $u_0$ & 0.2 \\
Safety gate temperature $T_q$ & 0.05 \\
Budget sigmoid temperature $T_{\mathrm{bud}}$ & 0.05 \\
Perf.-preserving margin $m_{\mathrm{pp}}$; Huber $\delta_H$ & 0.01;\enspace 0.10 \\
\bottomrule
\end{tabular*}
\caption{Hyperparameter configuration ($R=12$) used in the main experiments.}
\label{tab:hyperparams}
\end{table}
\FloatBarrier

\section{Additional experimental details}
\label{sec:additional-details}
\begingroup
\sloppy

\paragraph{MMLU subject split.}
The fixed split contains $34$ meta-train, $11$ meta-validation, and $12$ meta-test subjects. The meta-train subjects are:
{\small\ttfamily\raggedright
abstract\_algebra, anatomy, astronomy, business\_ethics,
clinical\_knowledge, college\_biology, college\_chemistry,
college\_computer\_science, college\_mathematics, college\_medicine,
computer\_security, conceptual\_physics, econometrics,
elementary\_mathematics, formal\_logic, high\_school\_biology,
high\_school\_chemistry, high\_school\_computer\_science,
high\_school\_geography, high\_school\_macroeconomics,
high\_school\_mathematics, high\_school\_physics,
high\_school\_statistics, human\_aging, international\_law,
logical\_fallacies, machine\_learning, management, marketing,
medical\_genetics, miscellaneous, nutrition, sociology,
us\_foreign\_policy.\par}
The meta-validation subjects are:
{\small\ttfamily\raggedright
college\_physics, electrical\_engineering, global\_facts,
high\_school\_european\_history,
high\_school\_government\_and\_politics,
high\_school\_microeconomics, high\_school\_psychology,
human\_sexuality, jurisprudence, moral\_disputes,
professional\_accounting.\par}
The meta-test subjects are:
{\small\ttfamily\raggedright
high\_school\_us\_history, high\_school\_world\_history, philosophy,
prehistory, moral\_scenarios, world\_religions, professional\_law,
professional\_medicine, professional\_psychology, public\_relations,
security\_studies, virology.\par}
The split is fixed across all sampling seeds.

\paragraph{C-Eval scopes.}
C-Eval-Full uses all $52$ subjects and contains $828$ labeled queries. Because public test labels are unavailable, all five development examples in each subject are used as support and up to $16$ labeled validation examples are used as queries. C-Eval-10 contains:
{\small\ttfamily\raggedright
middle\_school\_mathematics, middle\_school\_physics,
middle\_school\_politics, modern\_chinese\_history, plant\_protection,
professional\_tour\_guide, sports\_science, teacher\_qualification,
urban\_and\_rural\_planner, veterinary\_medicine.\par}
Every C-Eval-10 subject contributes $16$ queries. MetaNet receives no C-Eval parameter updates in either scope.

\paragraph{Adapted-baseline scope.}
All adapted rows preserve the DeepSeek backbone and original router. HyperRouter-style uses a support-conditioned linear head at fixed $k{=}6$; AdaMoE adapted uses the top-score margin; Dynamic MoE adapted uses a variable-$k$ rule inspired by Guo et al.~\citep{guo2025dynamic}; and Probe Pruning adapted uses a support-conditioned expert mask. These are inference-compatible adaptations, not exact reproductions of methods whose original versions require different training or pruning procedures; Probe Pruning latency therefore reflects a different execution path.
\endgroup

\section{Additional experimental results}

\begin{table}[!ht]
\centering
\footnotesize
\setlength{\tabcolsep}{2.5pt}
\begin{tabular*}{\linewidth}{@{\extracolsep{\fill}}lrrrr@{}}
\toprule
Method & Acc. & Act.\ ratio & Mean $k$ & Lat. (ms) \\
\midrule
\multicolumn{5}{l}{\textbf{MMLU: 12 subjects, 192 queries}} \\
Fixed $k{=}6$                 & 0.474 & 0.094 & 6.00 & 542.94 \\
Fixed $k{=}12$                & 0.495 & 0.188 & 12.00 & 557.65 \\
HyperRouter-style$^\ast$     & 0.406 & 0.094 & 6.00 & 539.32 \\
Probe Pruning adapted         & 0.188 & 0.094 & 6.00 & 180.45 \\
AdaMoE adapted$^\ast$         & 0.479 & 0.090 & 5.76 & 577.90 \\
Dynamic MoE adapted$^\ast$    & 0.453 & 0.094 & 6.00 & 572.18 \\
\method{} (aggressive)        & 0.438 & 0.036 & 2.28 & 516.52 \\
\method{} (conservative)      & 0.489 & 0.056 & 3.61 & 536.34 \\
\midrule
\multicolumn{5}{l}{\textbf{C-Eval-10: 10 subjects, 160 queries}} \\
Fixed $k{=}6$                 & 0.493 & 0.094 & 6.00 & 532.35 \\
Fixed $k{=}12$                & 0.581 & 0.188 & 12.00 & 531.36 \\
HyperRouter-style$^\ast$     & 0.443 & 0.094 & 6.00 & 487.80 \\
Probe Pruning adapted         & 0.281 & 0.094 & 6.00 & 146.15 \\
AdaMoE adapted$^\ast$         & 0.513 & 0.090 & 5.76 & 526.39 \\
Dynamic MoE adapted$^\ast$    & 0.519 & 0.094 & 6.00 & 541.96 \\
\method{}                     & 0.487 & 0.065 & 4.16 & 474.53 \\
\bottomrule
\end{tabular*}
\caption{Expanded comparison under the common inference harness. ``Adapted'' denotes an inference-compatible decision rule rather than an exact reproduction. C-Eval-10 is separate from C-Eval-Full in Table~\ref{tab:main-results}. $^\ast$ The original method requires training or replacement of routing components.}
\label{tab:expanded-baselines}
\end{table}
\FloatBarrier

\subsection{Operating-point sweep}

Table~\ref{tab:operating-points} reports four independently trained controller profiles. The \texttt{bt0p3}, \texttt{bt0p4}, \texttt{bt0p5}, and \texttt{bt0p8} tags denote nominal base budget targets $u_0=0.3$, $0.4$, $0.5$, and $0.8$, respectively, while the other shared settings follow Table~\ref{tab:hyperparams}. These targets configure the training objective; they are not achieved active ratios. We therefore report the realized metrics and do not claim that every row is Pareto-optimal.

\begin{table}[!ht]
\centering
\footnotesize
\setlength{\tabcolsep}{3.5pt}
\begin{tabular*}{\linewidth}{@{\extracolsep{\fill}}llrrrr@{}}
\toprule
Scope & Profile & Acc. & Act.\ ratio & Mean $k$ & Lat. (ms) \\
\midrule
\multirow{4}{*}{MMLU}
 & \texttt{bt0p3} & 0.460 & 0.027 & 1.75 & 527.07 \\
 & \texttt{bt0p4} & 0.472 & 0.031 & 1.98 & 527.96 \\
 & \texttt{bt0p5} & 0.477 & 0.040 & 2.55 & 516.02 \\
 & \texttt{bt0p8} & 0.494 & 0.045 & 2.88 & 502.15 \\
\midrule
\multirow{4}{*}{C-Eval-10}
 & \texttt{bt0p3} & 0.375 & 0.036 & 2.30 & 457.05 \\
 & \texttt{bt0p4} & 0.487 & 0.102 & 6.50 & 457.41 \\
 & \texttt{bt0p5} & 0.434 & 0.114 & 7.30 & 457.64 \\
 & \texttt{bt0p8} & 0.490 & 0.145 & 9.30 & 443.84 \\
\bottomrule
\end{tabular*}
\caption{Additional operating-point sweep. Active ratio is the realized mean top-$k$ divided by $64$. Latency is measured in the same three-GPU harness.}
\label{tab:operating-points}
\end{table}
\FloatBarrier

\subsection{Sensitivity to episode sampling}
\label{subsec:sampling-sensitivity}

The subject split is fixed in this check; the seed changes support/query sampling only. Each seed contains two episodes per subject and $384$ query predictions.

\begin{table}[!ht]
\centering
\footnotesize
\setlength{\tabcolsep}{2.5pt}
\begin{tabular*}{\linewidth}{@{\extracolsep{\fill}}lrrrrr@{}}
\toprule
Method & Seed & Acc. & Act.\ ratio & Mean $k$ & Lat. (ms) \\
\midrule
Fixed $k{=}6$              & 0 & 0.484 & 0.094 & 6.00  & 537.83 \\
Fixed $k{=}6$              & 1 & 0.526 & 0.094 & 6.00  & 538.22 \\
Fixed $k{=}12$             & 0 & 0.503 & 0.188 & 12.00 & 538.69 \\
Fixed $k{=}12$             & 1 & 0.531 & 0.188 & 12.00 & 540.36 \\
\method{} (conservative)   & 0 & 0.490 & 0.080 & 5.13  & 514.58 \\
\method{} (conservative)   & 1 & 0.521 & 0.074 & 4.75  & 511.83 \\
\method{} (aggressive)     & 0 & 0.461 & 0.070 & 4.47  & 535.72 \\
\method{} (aggressive)     & 1 & 0.492 & 0.060 & 3.82  & 496.27 \\
\bottomrule
\end{tabular*}
\caption{Repeated MMLU evaluation. Active ratio is mean top-$k$ divided by $64$.}
\label{tab:sampling-seeds}
\end{table}

Across the two seeds, accuracy is $0.505\pm0.030$ for fixed $k{=}6$, $0.517\pm0.020$ for fixed $k{=}12$, $0.506\pm0.022$ for conservative MetaNet, and $0.477\pm0.022$ for aggressive MetaNet (mean $\pm$ sample standard deviation). The corresponding mean top-$k$ values are $6.00\pm0.00$, $12.00\pm0.00$, $4.94\pm0.27$, and $4.15\pm0.46$. With only two seeds, these results measure sampling sensitivity rather than establish statistical significance.


\end{document}